\documentclass[journal]{IEEEtran}
\usepackage{hyperref}
\usepackage{titling} 
\usepackage{graphicx}
\usepackage{caption}

\usepackage{amsmath}
\usepackage{cite}
\usepackage{url}

\usepackage{graphicx}  
\usepackage{float}  

\begin{document}
\UseRawInputEncoding 
\bstctlcite{IEEEexample:BSTcontrol}
\title{\ MWaste: A Deep Learning Approach to Manage Household Waste\\ }

\author{Suman Kunwar\\
    \href{mailto:sumn2u@gmail.com}{sumn2u@gmail.com
    }\\}
    \date{}
        

\maketitle

\begin{abstract}
Computer vision methods have shown to be effective in classifying garbage into recycling categories for waste processing, existing methods are costly, imprecise, and unclear. To tackle this issue, we introduce MWaste, a mobile application that uses computer vision and deep learning techniques to classify waste materials as trash, plastic, paper, metal, glass or cardboard. Its effectiveness was tested on various neural network architectures and real-world images, achieving an average precision of 92\% on the test set. This app can help combat climate change by enabling efficient waste processing and reducing the generation of greenhouse gases caused by incorrect waste disposal.
\end{abstract}


\section{Introduction}

\IEEEPARstart{W}{aste} issue is a global concern and is on the rise due to the growth of urban areas and population, with predictions showing a potential increase of 70\% by 2050 if no measures are taken to address it \cite{kaza_what_2018}. The increasing complexity of waste composition and the absence of a standardized waste classification system make waste identification challenging, resulting in disparities in waste generation and management practices across different regions\cite{ferronato_waste_2019} \cite{shazia_2021}.

Comprehending household solid waste management practices is essential for the progress of integrated solid waste management\cite{fadhullah_household_2022}. Identifying waste plays a pivotal role in the waste management process as it enables facilities to manage, recycle, and diminish waste suitably, while ensuring compliance with regulations and monitoring their advancement over time.

Various studies and approaches that utilize deep learning models for efficient waste sorting and management, which can contribute to a more sustainable environment has been done. Models such as RWNet \cite{lin_applying_2022}, Garbage Classification Net \cite{liu_image_2022}, Faster Region-Based Convolutional Neural Network \cite{10000369}, and ConvoWaste \cite{Nafiz2023ConvoWasteAA} were proposed and evaluated for their high accuracy and precision rates in waste classification. These studies also highlight the importance of accurate waste disposal in fighting climate change and reducing greenhouse gas emissions. Some studies also incorporate IoT \cite{Cheema2022SmartWM} and waste grid segmentation mechanisms \cite{M2022TechnicalSF} to classify and segregate waste items in real-time.

By integrating machine learning models with mobile devices, waste management efforts can be made more precise, efficient, and effective. One of the research uses an app  that utilizes optimized deep learning techniques to instantly classify waste into trash, recycling, and compost with an accuracy of 0.881 on the test set \cite{narayan2021deepwaste}.

While it shows the potentiality the benchmarking with other state of art model is still needed and is limited in classifying waste into three types. In response, we introduce MWaste, a mobile app that utilizes computer vision and deep learning to classify waste materials into trash, plastic, paper, metal, glass, or cardboard types. The app provides users with suggestions on how to manage waste in a more straightforward and fun way.

The app is tested on various neural network architectures and real-world images, achieving highest precision of 92\% on the test set. This app can function with or without an internet connection and rewards users by mapping the carbon footprint of the waste they managed. The app's potential to facilitate efficient waste processing and minimize greenhouse gas emissions that arise from improper waste disposal can help combat climate change. Additionally, the app can furnish valuable data for tracking the waste managed and preserved carbon footprints.

The rest of this paper is structured as follows: Section II explains the system architecture of MWaste. Section III and IV detail the training and experimental evaluations. Finally, Section V summarizes the findings of this research.

\begin{figure*}
    \centering
    \includegraphics[width=1\textwidth]{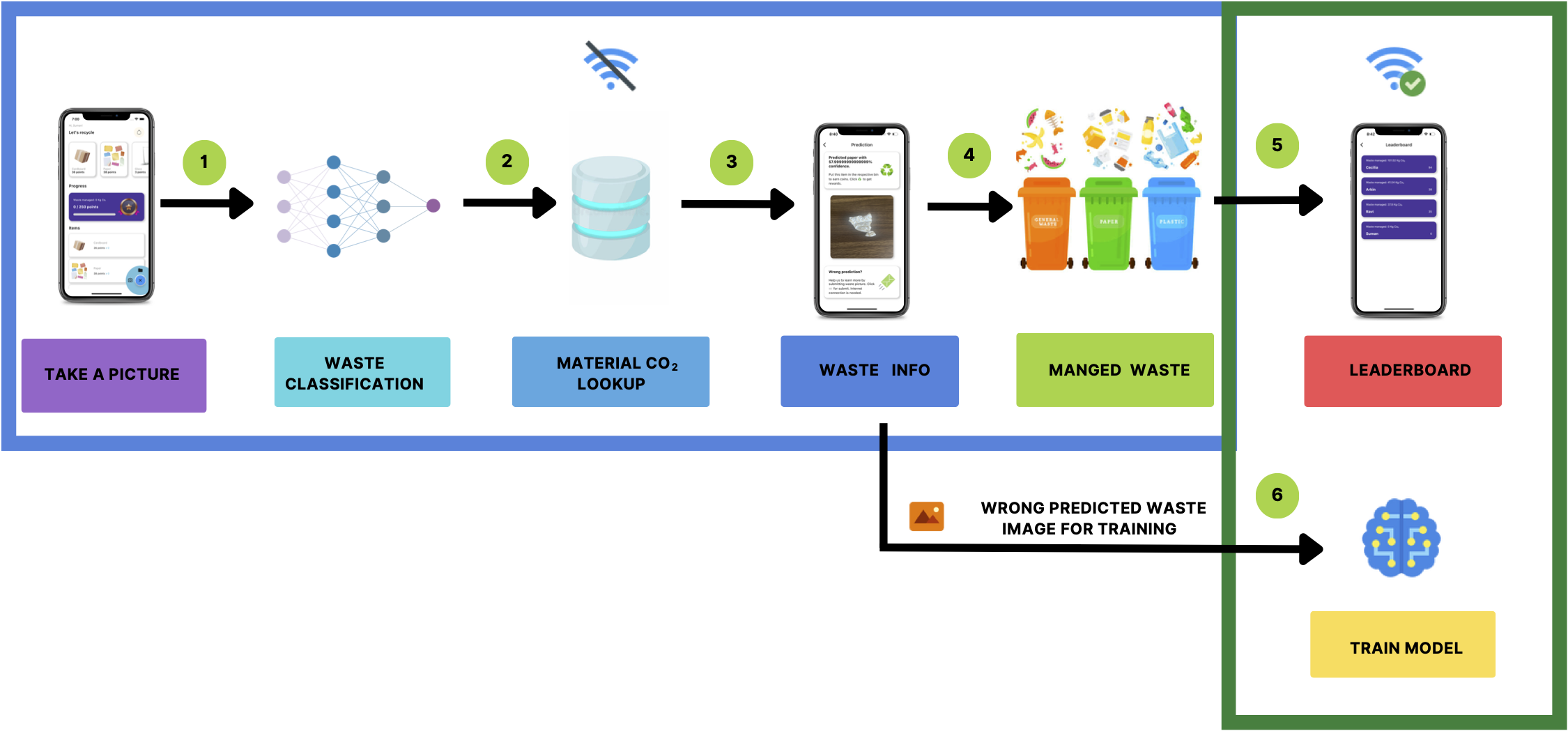}
    \caption{System Architecture of MWaste}
    \label{fig:system_arch}
\end{figure*}

\section{Methods}
This section discusses the architecture of the system and the flow of the process.

\subsection{System Architecture}

Classifying waste using machine learning is a challenging task since determining the recyclability or compostability of waste based on images is difficult due to the properties of the material being hard to detect from images. Besides, waste can take on various shapes and forms, which requires machine learning techniques to handle such variability and the recyclability of waste depends on the local recycling center's capabilities, which the app must consider.

Taking those considerations into account, the app is designed in such a way that feedbacks are collected from users and can operate smoothly with or without an internet connection. The waste image is obtained from the gallery or from camera, and is passed through the waste classification model, which is trained to categorize the waste. 

The classification model is the result of training a specific CNN model on a dataset of labeled images. Several state-of-the-art convolutional neural network methods is tested in this research, which included Inception V3 \cite{feng_office_2020}, MobileNet V2 \cite{yong_application_2023}, Inception Resnet V2 \cite{lee_novel_2021}, Resnet 50 \cite{10034869}, Mobile Net \cite{9699161}, and Xception \cite{9299017}.

The model is then converted into TensorFlow Lite model as they are highly optimized, efficient, and versatile, making them ideal for running real-time predictions on mobile \cite{noauthor_tensorflow_nodate}. Once identified, the model calculates the carbon emissions associated with the material and provides waste management recommendations. For misclassification, user can submit the waste image for further analysis. Managing waste earns reward points, and the amount of carbon footprint saved is also tracked.
An internet connection is required to submit wrongly predicted waste images and sync accumulated points.

\section{Training}
This section describes the training procedure and parameter settings used in this research.
\subsection{Datasets}

For this research, the publicly available trashnet dataset \cite{trashnet_dataset} is utilized, consisting of 2,527 images across six classes: glass, paper, cardboard, plastic, metal, and trash. These images were captured using Apple iPhone 7 Plus, Apple iPhone 5S, and Apple iPhone SE, with the objects placed on a white posterboard in sunlight or room lighting. The dataset was annotated by experts. To ensure robustness, 60\% of the images were used for training, 13\% for testing, and 17\% for validation.

\begin{figure*}
    \centering
    \includegraphics[width=0.8\textwidth,height=6cm]{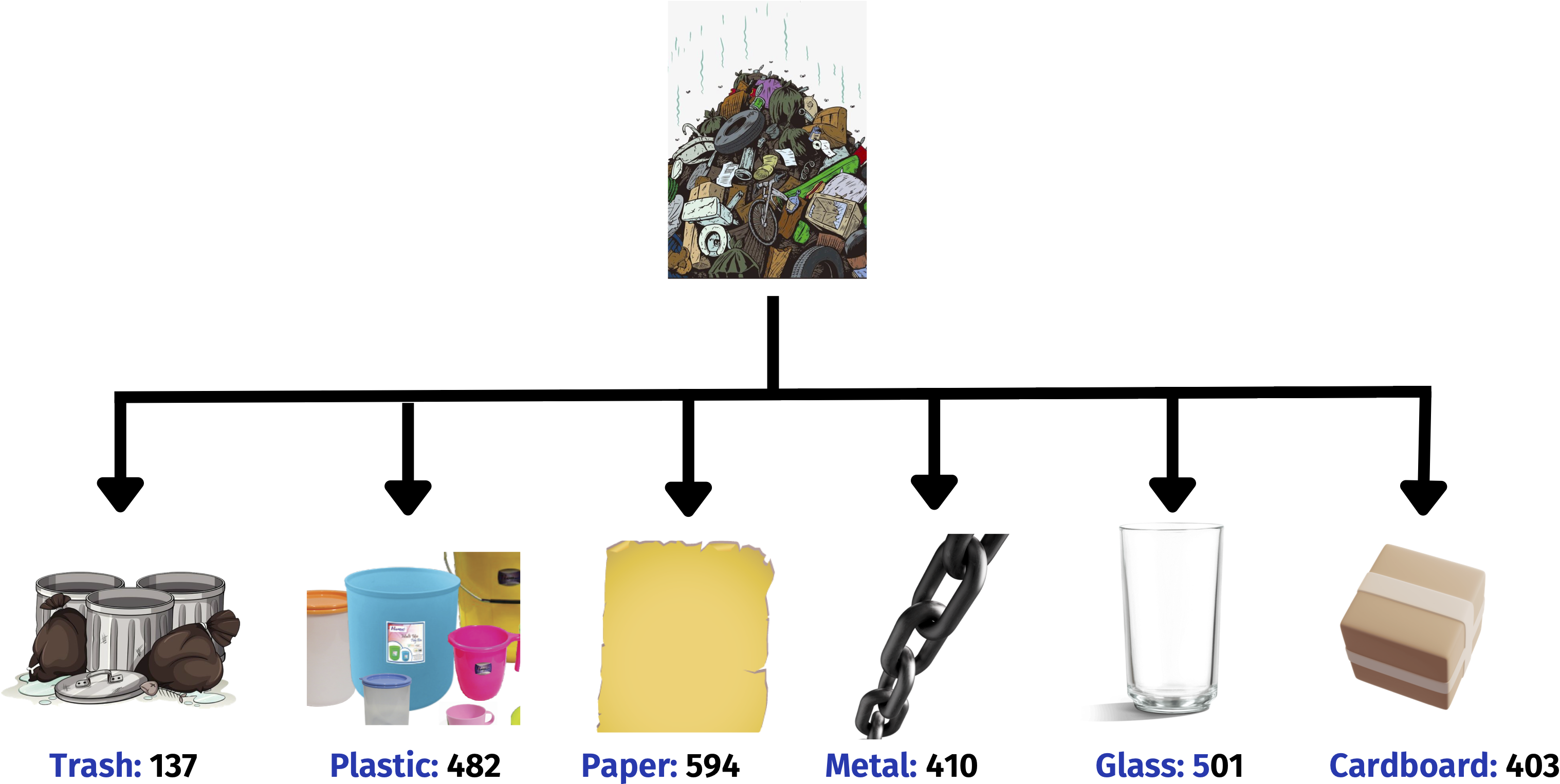}
    \caption{Datasets comprising the count of classes}
    \label{fig:img1}
\end{figure*}

\subsection{Procedure}
The gathered dataset is processed through different models while keeping all parameters constant. Subsequently, the outcomes are attentively analyzed. Categorical cross-entropy is employed to gauge the loss, as it is suitable for multiclass problems \cite{ho_real-world-weight_2020}. Meanwhile, accuracy serves as a metric, and Adam is the optimizer of choice, given that it applies momentum and adaptive gradient for computing adaptive learning rates for each parameter \cite{kingma_adam:_2017}.

Global average pooling is added to create one feature map per category in the final convolutional layer for the classification task \cite{ZHANG201836}. Three dense layers are then employed to learn complex functions and improve the accuracy of classification. To avoid overfitting, dropout is added as a regularization technique \cite{8682914}. Softmax is used as an activation function to convert the output values into probabilities \cite{zhou_mpce:_2019}.

\section{Evaluation}
In this section, different evaluation metrics are discussed and the results are compared based on them.
\subsection{Evaluation Metrics}
The evaluation measures can be used to explain the performance of various models. The study employs the Accuracy Score and F1 Score as evaluation metrics.

\subsubsection{Accuracy Score}
Classification accuracy is defined as the percentage of accurate predictions out of the total number of samples analyzed. To calculate accuracy in classification, the number of correct predictions is divided by the total number of predictions, and the resulting fraction is expressed as a percentage by multiplying it by 100\cite{li_evaluating_2019}. 
The formula for the accuracy score is as follows:
\begin{equation}\label{eq:relu_derivative}
Accuracy =  \frac{TP+TN}{TP+TN+FP+FN}
\end{equation}

\subsubsection{F1  Score}
When attempting to compare two models with contrasting levels of accuracy and recall, such as one with poor precision but strong recall, it can be challenging. Improving accuracy may have an adverse effect on recall, and vice versa, which can result in confusion \cite{uzen_surface_2021}. Hence, the F1-Score is utilized as a means of comparing the two sets and serves as a valuable metric for evaluating both recall and precision simultaneously.

The F1-Score is employed when dealing with imbalanced class data situations \cite{wardhani_cross-validation_2019}. As most real-world classification problems involve uneven case distributions, the F1-score is a more suitable metric for evaluating the model compared to accuracy.

\begin{equation}\label{eq:relu_derivative}
F1 = \frac{2*Precision*Recall}{Precision+Recall}
\end{equation}

\subsection{Model Evaluation}
Models are evaluated with same settings and their outputs are measured using evaluation metrics: accuracy score, and f1 -score. 

\begin{table*}[!b]\centering
    \captionsetup{justification=centering}
  \begin{center}
    \caption{Comparison of deep learning techniques and their accuracy for the given datasets}
    \label{tab:model_results_accuracy_comparison}
    \renewcommand{\arraystretch}{1.5}
    \begin{tabular}{l|c|c|c|c|c|c} 
      \textbf{} &\textbf{MobileNet} &\textbf{Inception V3} &\textbf{InceptionResNet V2}&\textbf{ResNet 50}&\textbf{MobileNet V2}&\textbf{Xception}\\
      \hline
      \hline Accuracy & 0.87 & 0.89 & 0.92 & 0.84 & 0.86 & 0.92 \\
    \end{tabular}
  \end{center}
\end{table*}

\begin{table*}[!b]\centering
    \captionsetup{justification=centering}
  \begin{center}
    \caption{Comparison of deep learning techniques based on their F1-scores for each class in the given datasets}
    \label{tab:model_results_comparison}
    \renewcommand{\arraystretch}{1.5}
    \begin{tabular}{l|c|c|c|c|c|c} 
      \textbf{F1-score} &\textbf{MobileNet} &\textbf{Inception V3} &\textbf{InceptionResNet V2}&\textbf{ResNet 50}&\textbf{MobileNet V2}&\textbf{Xception}\\
      \hline
      \hline Cardboard & 0.94 & 0.95 & 0.97 & 0.91 & 0.97 & 0.96 \\
      \hline Glass & 0.85 & 0.86 & 0.90 & 0.86 & 0.78 & 0.91  \\
      \hline Metal & 0.86 & 0.88 & 0.91 & 0.83 & 0.86 & 0.95 \\
       \hline Paper & 0.91 & 0.92 & 0.96 & 0.86 & 0.93 & 0.94 \\
      \hline Plastic & 0.89 & 0.88 & 0.91 & 0.86 & 0.83 & 0.90 \\
      \hline Trash & 0.52 & 0.68 & 0.68 & 0.41 & 0.32 & 0.67 \\
    \end{tabular}
  \end{center}
\end{table*}

\begin{figure}[H]
\centering
\captionsetup{justification=centering}
\includegraphics[width=0.50\textwidth]{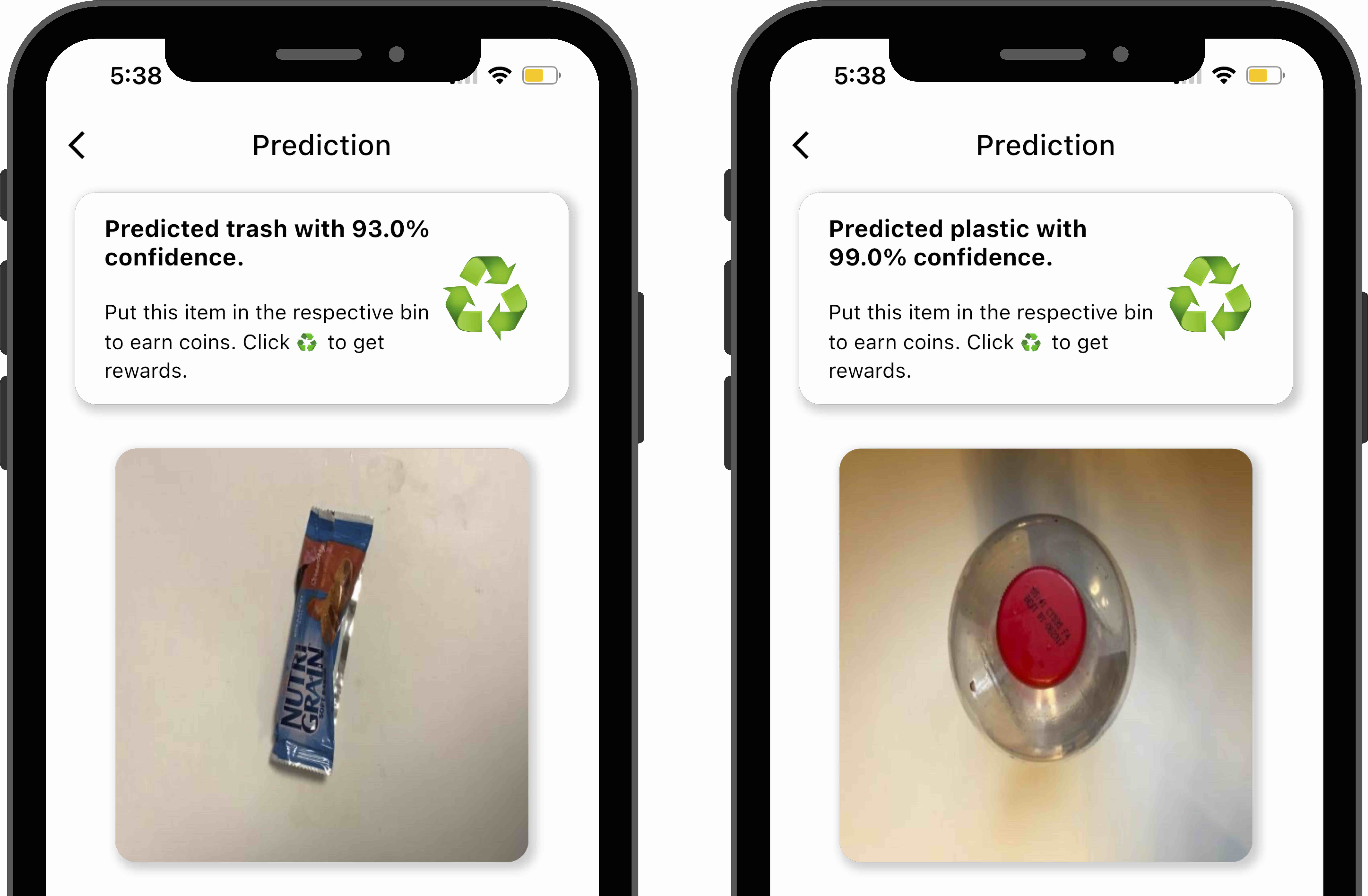}
\caption{ Classification output in MWaste App}
 \label{fig:predicted_results}
\end{figure}

After comparing the models as shown in \autoref{tab:model_results_accuracy_comparison}, it can be seen that InceptionResNetV2 and Xception have higher accuracy, but the loss is higher for InceptionResNetV2 and Inception V3 models. \autoref{fig:predicted_results} illustrates the classification result of a waste material from the given training set.

\begin{figure}[H]
\centering
\captionsetup{justification=centering}
\includegraphics[width=0.50\textwidth, height=15.5cm]{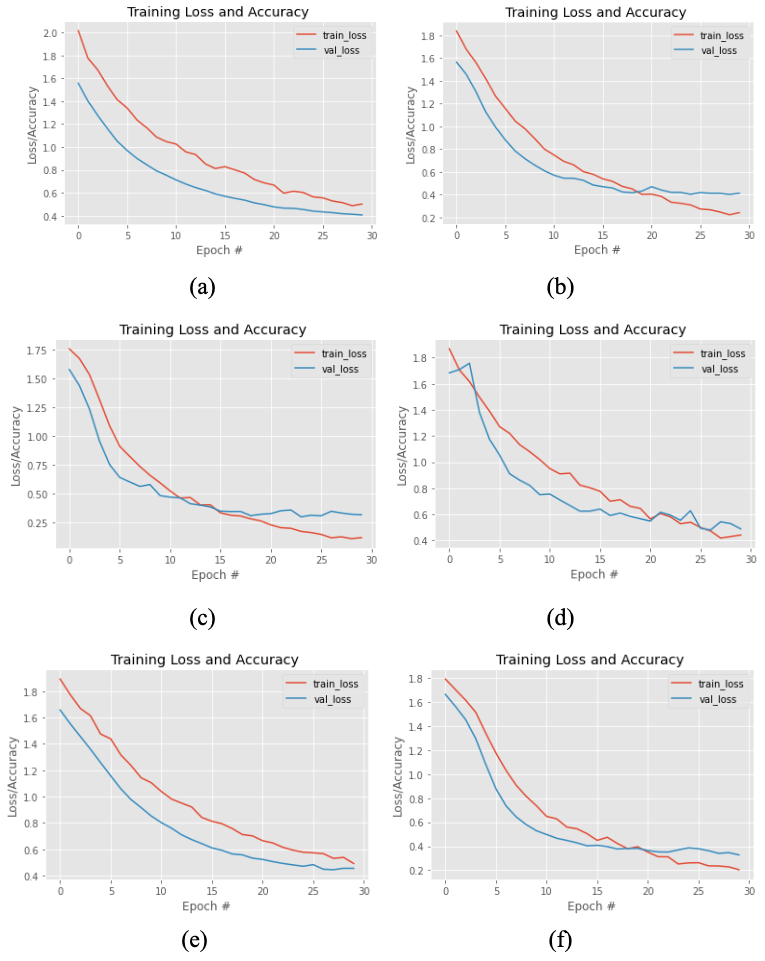}
\caption{ Training Loss and Accuracy graph of (a) MobileNet (b) Inception V3 (c) InceptionResNet V2 (d) ResNet50 (e) MobileNet V2 (f) Xception models with given datasets}
 \label{fig:accuracy-loss-results}
\end{figure}

 Accuracy and loss of each model during training is shown in \autoref{fig:accuracy-loss-results}. 
\vspace{1em}

\section{Conclusion and Future Work}

This study presents a mobile application that utilizes deep learning techniques to classify waste in real-time. The app categorizes waste into six groups, including plastic, paper, metal, glass, cardboard, and trash, and is publicly available with a trained model\footnote{\url{https://github.com/sumn2u/deep-waste-app}}. The app incorporates gamification strategies, such as a leaderboard based on waste management points, to motivate users to dispose of waste properly.

The team plans to improve the accuracy of the classification system, form partnerships with local recycling companies, and expand the dataset to raise awareness of environmental impacts and reduce incorrect waste disposal.

\section{Acknowledgements}
My heartfelt appreciation goes out to Gary Thung and Mindy Yang for sharing the TrashNet dataset on Github for public use. This dataset has proven to be an invaluable asset for my research or project on waste management and classification, and I am deeply thankful for their hard work in gathering and disseminating this information to a larger audience.

\bibliographystyle{IEEEtran}

\bibliography{references}

\newpage
\onecolumn

\end{document}